# A Spiking Neural Network Structure Implementing Reinforcement Learning

**Mikhail Kiselev**
Chuvash State University
Cheboxary, Russia
`mkiselev@chuvsu.ru`

## Abstract

At present, implementation of learning mechanisms in spiking neural networks (SNN) cannot be considered as a solved scientific problem despite plenty of SNN learning algorithms proposed. It is also true for SNN implementation of reinforcement learning (RL), while RL is especially important for SNNs because of its close relationship to the domains most promising from the viewpoint of SNN application such as robotics. In the present paper, I describe an SNN structure which, seemingly, can be used in wide range of RL tasks. The distinctive feature of my approach is usage of only the spike forms of all signals involved – sensory input streams, output signals sent to actuators and reward/punishment signals. Besides that, selecting the neuron/plasticity models, I was guided by the requirement that they should be easily implemented on modern neurochips. The SNN structure considered in the paper includes spiking neurons described by a generalization of the LIFAT (leaky integrate-and-fire neuron with adaptive threshold) model and a simple spike timing dependent synaptic plasticity model (a generalization of dopamine-modulated plasticity). My concept is based on very general assumptions about RL task characteristics and has no visible limitations on its applicability. To test it, I selected a simple but non-trivial task of training the network to keep a chaotically moving light spot in the view field of an emulated DVS camera. Successful solution of this RL problem by the SNN described can be considered as evidence in favor of efficiency of my approach.

## 1 Introduction and motivation

Nowadays, lack of efficient and sufficiently general learning algorithm implementations for spiking neural networks (SNN) is one of major obstacles for massive penetration of SNN-based technologies in all spheres of life. Advance in learning algorithm implementation is different in different classes of learning tasks. For example, implementation of unsupervised learning in SNN is relatively more developed technique, perhaps, due to the fact that the classic synaptic plasticity model STDP (spike timing dependent plasticity) is most suitable for unsupervised learning (as it is shown in [1-3]). Situation in the field of supervised learning and, especially, reinforcement learning (RL) [4] is worse. Although SNNs used to solve RL problems have been considered in a number of works (e.g. [5-8]), they often represent SNN as a functional equivalent of traditional neural networks due to use of firing rate - based information coding, firing rate - based plasticity rules and reward signal available at every moment as a numeric value (often called in neurophysiological terms - *dopamine concentration* [9]).

For example, in [5], it is demonstrated that simple RL problems can be solved by stochastic leaky integrate-and-fire (LIF) neurons with STDP plasticity modulated by the current value of reward which may be positive or negative. The network consisting of layer of LIF neurons with firing rate – based plasticity rules including current and predicted dopamine concentrations and chaotically interconnected neurons with rate – based unsupervised plasticity of their connections with input nodes was considered in [6]. It was shown that this network achieves good results on public RL benchmarks like the Mountain Car problem [16]. RL tasks from the same benchmark set (OpenAI Gym) were used to test dopamine – modulated rate – based plasticity rules proposed in [7]. In the paper [8], the reward – modulated plasticity rules similar to the rules from [5] were extended to inhibitory synapses. The technique for converting deep learning networks solving RL problems to SNNs was developed, in particular, in [17]. SNN in the form of a liquid state machine (LSM) plays an important role in the approach to RL proposed in [18] however, in this work, LSM neurons are not plastic and are not involved in the learning process – they just form high dimensional representation of the current environment dynamics.

In contrast to these and other similar works, the aim of this study is the "native" SNN approach to RL where all involved signals have the spiking nature and exact firing times are important. The first factor became crucial with appearance of dedicated neurochips emulating SNN like TrueNorth [10] and Loihi [11]. Many of them operate with spikes only – exchange of numeric signals or global access to a numeric values imitating dopamine concentration are not implemented in them. It is another novel aspect of my study – the requirement that all features of my network/plasticity models would permit efficient implementation on the modern SNN-emulating neurochips (I took Loihi as an example).

Let us formalize the problem – what is meant by the SNN-native RL? First of all, it is stipulated that all relevant information should have the spike form. Namely, not only the input sensory data should be represented as spike streams but network's effectors also should interpret spikes from certain neurons as commands to execute and reward/punishment signals should have the spike nature as well. Second, all the signals are mutually asynchronous. There are no framed "examples" - only continuous data flows instead. The reward/punishment signals are asynchronous, too, and may arrive with significant and a priori unknown delay with respect to the network actions evaluated.

Thus, SNN interacts with its environment via three channels: it receives information about current environment state, it produces control signals changing environment state in some way, and it receives evaluation of its commands in form of reward and punishment signals which may be generated after these actions with unknown delay. The aim of RL is to organize SNN behavior so that number of reward signals would be maximized while number of punishment signals would be minimized.

In the present work, this goal is achieved due to the special selection of SNN structure, the slightly generalized version of the frequently used LIFAT (leaky integrate-and-fire neuron with adaptive threshold [12]) neuron model and four types of synapses (excitatory, inhibitory, plasticity inducing and activating). All these features are described in the next Section.

## 2    Neuron, Network, Plasticity

The overall network structure is depicted on Figure 1. At first, let us consider it in general to convey the main idea of this work and, after that, describe it more formally.

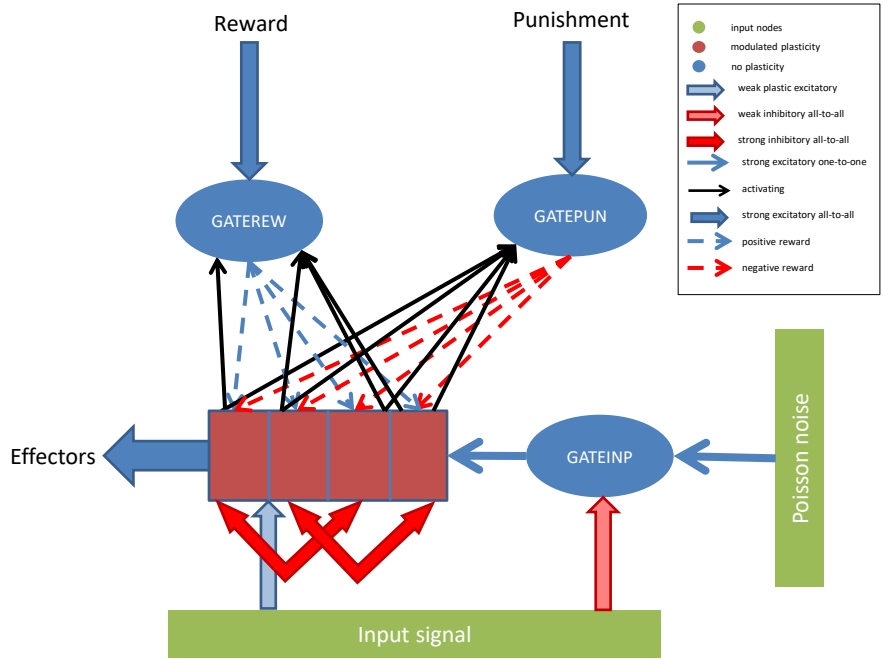

Figure 1. The schematic structure of an SNN implementing RL – see the detailed description in the text.

The external signals carrying information about environment have the form of spike trains. These signals are sent to the learning neurons (colored brown on Figure 1) via plastic synapses. Each learning neuron is connected to some random subset of input nodes. At the beginning weights of all these connections are equal. The learning neurons form several groups depicted on Figure 1 as sections of the brown rectangle. Every group is responsible for a certain action interpreted by the effectors and impacting the environment. Strength of this action is coded by the amount of active (firing) neurons in the group. If some actions are antagonistic by their nature or are mutually excluding then the respective groups are interconnected by strong inhibitory links (in the all-to-all manner) – see the thick red arrows below the brown rectangle.

In the SNN described, the only plastic connections are the connections between input nodes and learning neurons. Plasticity of these afferent synapses is governed by the neuronal populations GATEREW and GATEPUN in the following way. The number of neurons in the populations GATEREW and GATEPUN equals to number of groups of learning neurons. All learning neurons in every group have projections to the corresponding neuron in GATEREW and GATEPUN populations via the so called *activating* synapses. The basic state of neurons in GATEREW and GATEPUN is *inactive*. In this state, they are unable to react to any stimulation – they are unable to fire. Upon receiving a spike at its activating synapse, such a neuron becomes active i.e. becomes a normal neuron whose membrane potential is changed by incoming excitatory and inhibitory spikes and which can fire etc. Duration of this active state is proportional (or, simply speaking, equal) to the weight of the activating synapse. Neuron is active while activity period triggered by any of its activating synapses lasts. The GATEREW/GATEPUN neurons are fed by the reward or punishment signals, respectively. The excitatory synapses used to transfer these signals to GATEREW and GATEPUN are so strong that force the neurons to fire unless they are inactive. Thus, the GATEREW/GATEPUN neurons fire if 1) they have been activated by spikes from learning neurons in the recent past **and** 2) they receive reward/punishment spikes. Therefore, the GATEREW/GATEPUN neurons can be utilized to approve or condemn a recent action initiated by the network. They do it using the *plasticity inducing* connections from them to learning neurons. When a neuron receives spike on its plasticity inducing synapse, all its plastic synapses having obtained spike in the recent past (during the time interval of length $\tau_P$) change their weights by the weight of this plasticity inducing synapse. Depending on sign of this weight, they may be potentiated or depressed. The *plasticity period* $\tau_P$ is a constant of the neuron model. Each GATEREW/GATEPUN neuron is connected via plasticity inducing synapses with all learning neurons in its group.

It is quite obvious that in case of the appropriate choice of model parameters, the correct pairs <environment state, action> will be rewarded (due to facilitation of the respective synapses) while the harmful pairs will be suppressed. I hypothesized that this scheme would need a significant addition. Indeed, if the input signal is weak and/or the plastic synapses are still not tuned then learning neurons do not fire (or fire very infrequently), the network does not issue commands to effectors, does not activate the gates and, therefore, cannot learn anything. In order to facilitate learning on the early stages, I introduced the following mechanism. I added to the structure described above the sources of random Poisson spiking signal. Their quantity is equal to number of learning neurons. These noise sources are one-to-one connected to neurons in the population GATEINP via strong excitatory synapses (forcing these neurons to fire unless they are strongly inhibited). At the same time, the GATEINP neurons are inhibited by input signal. If input signal is weak, this inhibition does not work and the Poisson noise forces the learning neurons to fire (and, therefore, to learn). If input signal is strong then the network uses input signal for learning. In this case, the Poisson noise is blocked – to prevent noise influencing the learning process. While, as we will see later, this mechanism was found to be unnecessary in the RL problem solved in this study, I still think that it may be useful in other RL tasks.

Let us consider the models of neuron and synaptic plasticity in more detail. In this research, I use the so called LIFAT (leaky integrate-and-fire neuron with adaptive threshold [12]) model. It is a popular neuron model often used in SNN research. Besides that, it is efficiently implementable on the modern neurochips (e.g. Loihi).

The simplest current-based delta synapse model is used for all excitatory and inhibitory synapses. It means that every time the synapse receives a spike, it instantly increases the membrane potential by the value of the synaptic weight (positive or negative – depending on the synapse type). Thus, the state of a neuron at the moment $t$ is described by its membrane potential $u(t)$ and its threshold potential $u_{THR}(t)$. Dynamics of these values are defined by the equations

$$\begin{cases} \frac{du}{dt} = -\frac{u}{\tau_v} + \sum_{i,j} w_i^+ \delta(t - t_{ij}^+) - \sum_{i,j} w_i^- \delta(t - t_{ij}^-) \\ \frac{du_{THR}}{dt} = -a \, \text{sgn}(u_{THR} - 1) + \sum_k \hat{T} \delta(t - \hat{t}_k) \end{cases} \qquad (1)$$

and the condition that if $u$ exceeds $u_{THR}$ then the neuron fires and value of $u$ is reset to 0. It should be noted that for sake of simplicity all potentials are rescaled so that after the long absence of presynaptic spikes $u \to 0$ and $u_{THR} \to 1$. The meaning of the other symbols in (1) is the following: $\tau_v$ – the membrane leakage time constant; $a$ – the speed of decreasing $u_{THR}$ to its base value 1; $w_i^+$ - the weight of $i$-th excitatory synapse; $w_i^-$ - the weight of $i$-th inhibitory synapse; $t_{ij}^+$ - the time moment when $i$-th excitatory synapse received $j$-th spike; $t_{ij}^-$ - the time moment when $i$-th inhibitory synapse received $j$-th spike; $\hat{T}$ – $u_{THR}$ is incremented by this value when the neuron fires at the moment $\hat{t}_k$.

As it was said above, I introduced only one non-standard feature in the LIFAT model (for neurons in the GATEREW and GATEPUN populations) – the ability to sleep. In the sleeping mode, presynaptic spikes do not change value of the membrane potential. The neuron is activated (and becomes a normal LIFAT neuron) by spikes arriving at its special activating synapses. Such a synapse keeps the neuron in the active state for period of time equal to its weight. The neuron is active while at least one of its activating synapses keeps it in active state. Obviously, this feature can be easily implemented in hardware.

The synaptic plasticity model is also very simple. Its main distinctive feature is the same as in our previous research works [13-15]. Namely, synaptic plasticity rules are applied to the so called *synaptic resource W* instead of the synaptic weight $w$. There is functional dependence between $W$ and $w$ expressed by the formula

$$w = w_{\min} + \frac{(w_{\max} - w_{\min}) \max(W, 0)}{w_{\max} - w_{\min} + \max(W, 0)}, \qquad (2)$$

where $w_{min}$ and $w_{max}$ are constant (in this study, $w_{min} = 0$).

In this model, the weight values lay inside the range [$w_{min}$, $w_{max}$) - while $W$ runs from $-\infty$ to $+\infty$, $w$ runs from $w_{min}$ to $w_{max}$. When $W$ is either negative or highly positive, synaptic plasticity does not affect a synapse's strength. Instead, it affects its *stability* – how many times the synapse should be potentiated or depressed to move it from the saturated state. Thus, to destroy the trained network state, it is necessary to present the number of "bad" examples close to the number of "good" examples used to train it. Such an approach makes the weight values naturally limited and solves the hard problem of catastrophic forgetting.

The plasticity rule proposed in this study is very simple. The plastic neurons have special plasticity inducing synapses. When a plasticity inducing synapse obtains spike, synaptic resources of all plastic synapses of this neuron having received at least one presynaptic spike during the time interval of the length $\tau_P$ in the past are modified by the value equal to weight of this plasticity inducing synapse (it may be positive or negative).

In order to prevent unlimited potentiation or depression of neuron's synapses, I added one more rule - constancy of neuron's total synaptic resource. When some synapse is potentiated or depressed, the resources of all other plastic synapses are modified in the opposite direction by the same value calculated from the condition that the total synaptic resource of the neuron should remain the same. It should be noted that for purpose of emulation speed up, this weight renormalization procedure is not performed at every synaptic weight modification. Instead, it is invoked only when the accumulated total synaptic resource change becomes sufficiently great.

# 3    The Test RL Task – Tracking Light Spot by Imitated DVS Camera.

In order to test capability of solving RL problems by the SNN proposed I selected the following simple but non-trivial task. I emulated a square area of size 2×2 m with black background inside which a light spot is moving chaotically. A square region of size 1×1 m inside this area is observed by an imaginary camera. This region can be moved over the whole area by signals from the SNN. The task is to track the light spot by this camera keeping it near the center of camera's view field. This camera behaves like a DVS camera because it sends spiking signal characterizing current brightness of each pixel together with pixel brightness dynamics. The camera has the view field of the size 20×20 pixels. 3 spike streams (channels) are associated with each pixel. Spike frequency in the first channel is proportional to the current pixel brightness. The other two channels send spikes every time the pixel brightness increases or decreases by a certain value. Thus, the whole input signal includes 1200 spiking channels. The thresholds used to convert brightness and brightness changes into spikes are selected so that the mean spike frequency in all channels would be close to 30Hz. Using these three channels, the network can obtain information about static and moving objects.

The whole emulation takes 1000000 time steps (we assume that 1 time step = 1 msec). The time necessary for the light spot to cross the whole square area is several hundred milliseconds. It is assumed that during first 800 sec the network is trained. Last 200 sec are used to evaluate the learning results. The exact criterion is the total time when the distance $d$ between the centers of the light spot and the camera view field is less than 0.15 m.

Learning neurons in the SNN form 4 equal size groups whose spikes are interpreted as commands to move camera in four directions (up, right, down, left). Size of these groups is one of the network parameters optimized by genetic algorithm described below. Since two pairs of these groups act antagonistically, the corresponding groups are mutually connected by the strong inhibitory links. Every command spike increases camera velocity in the respective direction by 0.01 m/msec (10 m/sec). The medium, which the camera moves in, is viscous – the moving camera is slowed down with the constant speed 0.03 m/msec (30 m/sec).

The punishment/reward signal is generated by the following algorithm. At the emulation beginning, the value of $d$ is stored. Every time $d$ is increased by 0.1m, a punishment spike is emitted and the new value of $d$ is stored. When $d$ becomes 0.1m less than the stored value, a reward spike is emitted and $d$ is stored. Besides that, reward spikes are emitted with 6 msec period while $d < 0.15$.

The code emulating the environment described can be found at https://github.com/MikeKis/AShWin (the branch `ArXiv1`).

In order to find the best SNN parameters for solution of this problem I used an optimization procedure based on genetic algorithm. To diminish the probability of accidental bad or good result, I averaged the criterion values for 3 tests with the same SNN parameters but with different light spot trajectories. The population size was 300, mutation probability per individual equaled to 0.5, elitism level was 0.1.

Figure 2 shows the results obtained by genetic algorithm. We see that near-optimum network is found very quickly.

The video showing how the SNN learns to catch the light spot can be downloaded from https://disk.yandex.ru/d/YNu1nYXB6eLhmw. We see that SNN learns to track the light spot quickly (during first 50 sec) and does it very well. It should be noted that the criterion value cannot be close to 1 because sometimes the spot closely approaches the square area boundaries where it cannot be caught in the camera's view field center. As it is shown on Figure 3 depicting learning curves for 3 launches of the best SNN emulation, the network is reliably trained to catch the spot.

The parameters varied in the optimization procedure, the ranges of their variation and their optimum values found are presented in Table 1.

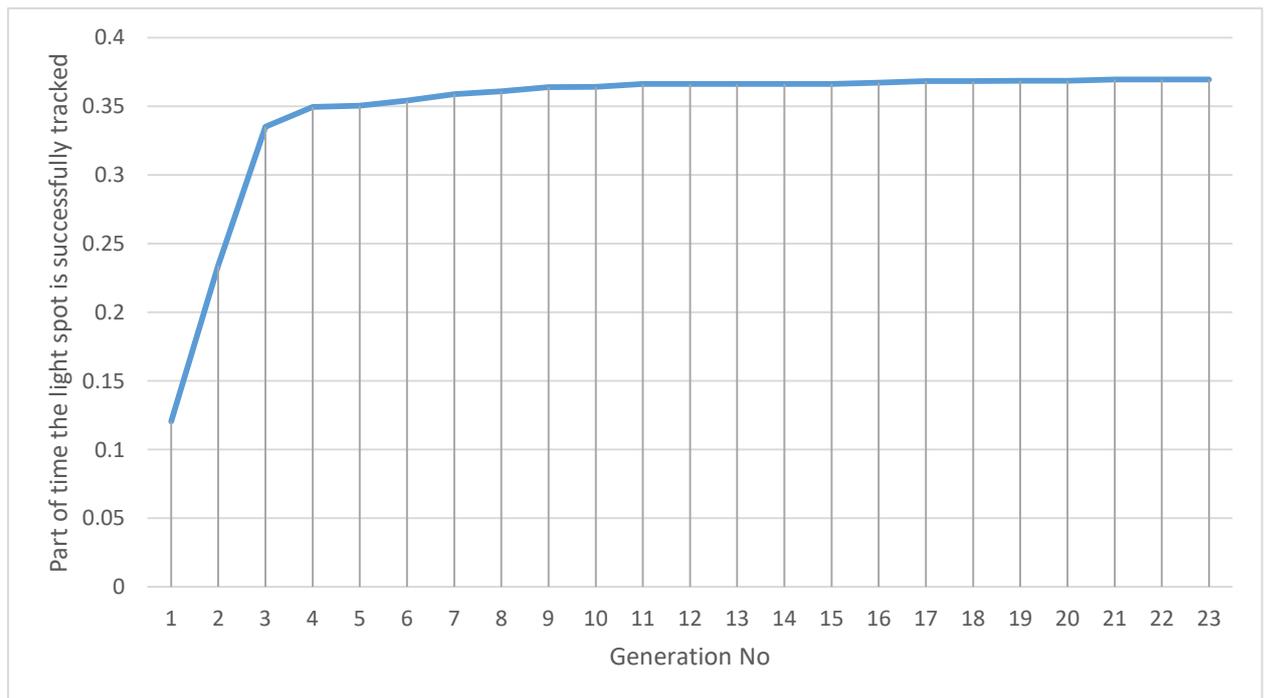

Figure 2. The course of maximization of the light spot successful tracking time in sequential generations of genetic algorithm.

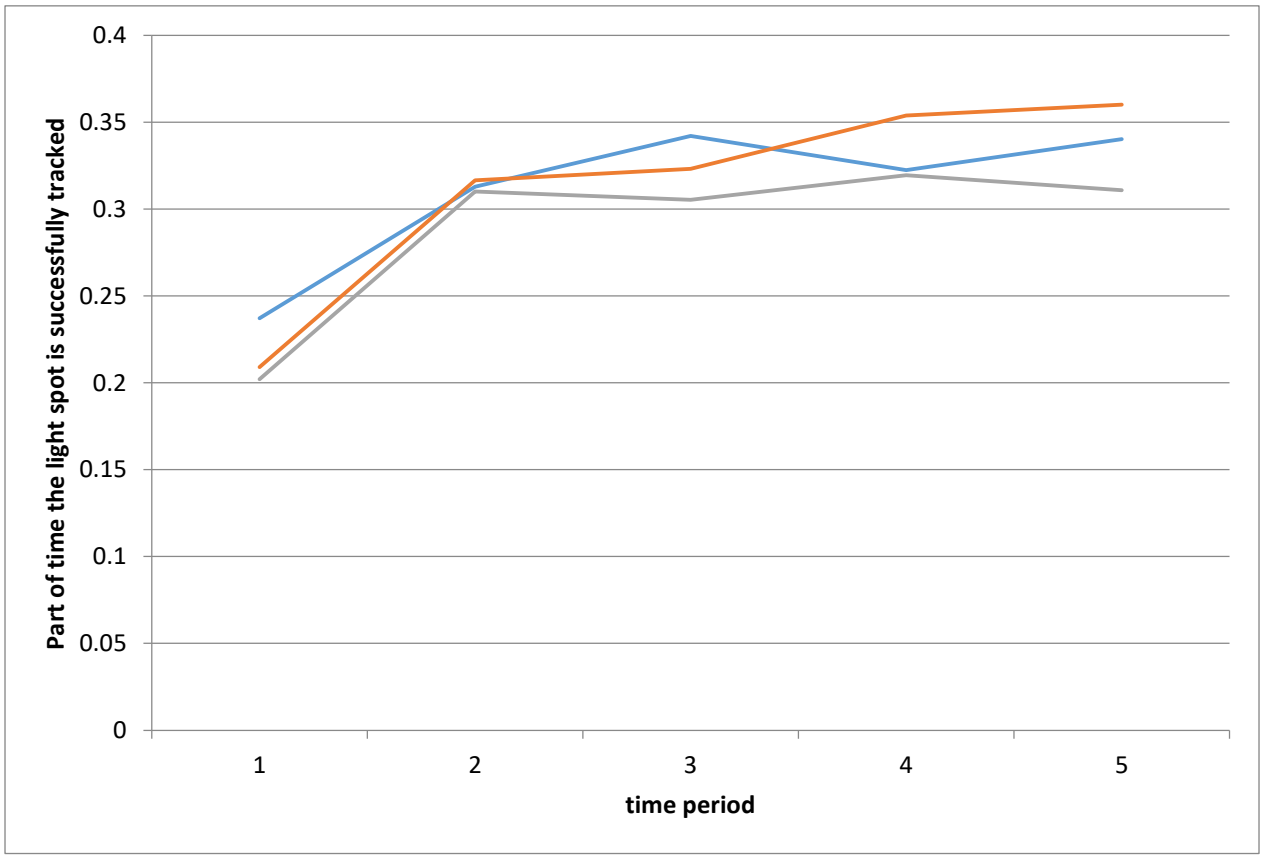

Figure 3. The learning curves for 3 launches of the best SNN emulation for 5 200-sec periods of emulation.

**Table 1. The SNN parameters optimized.**

| Parameter optimized | Value range | Optimum value |
| --- | --- | --- |
| Number of neurons per movement direction | 1-100 | 10 |
| Input Poisson noise frequency | 0.0001Hz – 10 Hz | 0.001Hz (= no noise, in fact) |
| Maximum afferent weight $w_{max}$ | 0.12 – 1.2 (if the base value of the threshold membrane potential equals to 1) | 0.638 |
| Initial value of afferent synaptic resource | 0.3 - 3 | 2.66 |
| Share of input nodes connected to one learning neuron | 0.01 – 0.8 | 0.77 |
| Weight of inhibitory connections of input nodes with GATEINP neurons | 0.12 – 1.2 | 0.13 (not important) |
| Membrane potential decay time of learning neurons $\tau_v$ | 1 msec – 30 msec | 2 msec |
| Ratio of $\hat{T}$ and $a$ (see (1)) for learning neurons (it corresponds to equilibrium time interval between consecutive postsynaptic spikes) | 1 msec – 100 msec | 17 msec |

| Threshold potential increment of learning neurons $\hat{T}$ | 0.12 - 12 | 10 |
|---|---|---|
| Plasticity time $\tau_P$ | 3 msec – 100 msec | 18 msec |
| Membrane potential decay time of neurons from GATEREW and GATEPUN | 1 msec – 30 msec | 8 msec |
| Ratio of $\hat{T}$ and $a$ (see (1)) for GATEREW/GATEPUN neurons (it corresponds to equilibrium time interval between consecutive postsynaptic spikes) | 3 msec – 100 msec | 41 msec |
| Threshold potential increment of GATEREW/GATEPUN neurons $\hat{T}$ | 0.12 - 36 | 8 |
| Membrane potential decay time of neurons from GATEINP | 3 msec – 30 msec | 14 msec (not important) |
| Weight of activating synapses | 3 msec – 300 msec | 4 msec |
| Ratio of absolute values of negative and positive plasticity inducing synapse weights (all positive plasticity inducing synapse weights are equal to 0.12) | 0.3 - 30 | 26.6 |

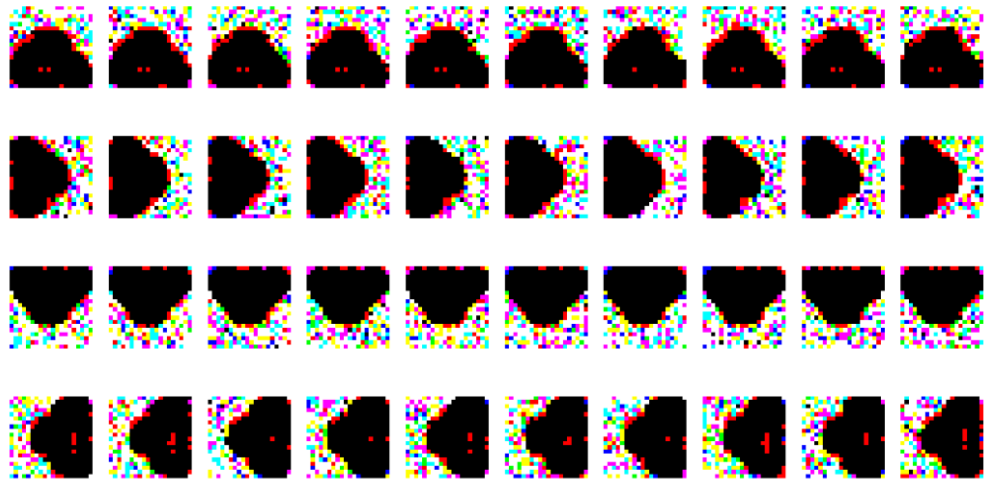

Figure 4. The learnt weights of neurons in the best SNN at the simulation end. Only strong connections are depicted. Each square corresponds to the camera view field. The pixel color codes input channel (red – current brightness, green – brightness increase, blue – brightness decrease).

The weights learnt in this task are depicted on Figure 4. Each square corresponds to one learning neuron. Each row corresponds to one camera movement direction in the order "up, right, down, left". Squares are broken to pixels corresponding to camera's input pixels. Shown pixel color codes strong ($W > 30$) afferent connections. Red color codes connections with brightness channel (first 400 input nodes); green codes strong connections with input nodes reacting to pixel brightness increase (next 400 input nodes); at last, blue corresponds to pixel brightness decrease nodes (last 400 input nodes). If several connections with the same pixel are strong then it is displayed by a mixed color. For example, if some pixel has strong connections in all three channels then the corresponding pixel on Figure 4 is colored white. We see that distribution of strong connections matches camera movement direction. It is interesting that at the boundaries of the colored zones, only connections to the input nodes reacting to current brightness are strong. Noticeable, all neurons have strong connections with few brightness channel nodes in the opposite directions as well. Probably, it is not accidental but at present I cannot explain this observation.

Let us discuss and summarize the results obtained.

- In this task, the noise injecting mechanism appears to be unnecessary (therefore, the SNN does not need the Poisson input nodes and the GATEINP population). Nevertheless, I still believe that this mechanism may be useful in other RL tasks.
- The SNN solving this problem efficiently and reliably is surprisingly small – it includes only 48 neurons.
- Connections between learning neurons and input nodes are almost "all-to-all". However they should not be exactly "all-to-all" because in this case all learning neurons would be identical (since the neuron model is deterministic and all initial weights are equal). Such a network shows more than 2 times worse result. I expected that learning neurons should be strongly mutually independent, however, it was found to be unnecessary – although the neurons have many common inputs they are (or become) sufficiently independent to make the learning process stable and to vary the controlling signal strength using population coding.
- Relative effect of punishment spikes is unexpectedly great. It may be explained by the reward signal generation algorithm used – where the considerable part of reward spikes is not related to actions but rather to the current camera position (when the spot is almost at the camera view field center).

# 4   Conclusion.

In this paper, I tried to solve the problem of reinforcement learning in its purely spiking formulation – when all signals involved in the learning process – sensory signal, commands sent by network and reward/punishment signal – are represented in the form of spike trains. It is especially important from view point of application of modern neurochips where all data are represented in form of spikes transmitted as AER packets and global variables expressing punishment/reward levels cannot be easily implemented. In general, efficient implementation of synapse/neuron/network models on this kind of neurochips was keynote of this study.

Pursuing this goal, I designed a very simple SNN architecture, a modification of LIFAT neuron model and a simple variant of dopamine-modulated spike timing dependent plasticity. To evaluate efficiency of this SNN in RL tasks, I utilized an easily controlled but non-trivial environment where emulated DVS camera should learn to track chaotically moving light spot. In order to find the optimum SNN configuration, the SNN parameter optimization procedure based on genetic algorithms was designed. Using this procedure, a small and efficient SNN solving this problem has been found. Some of its properties were unexpected and gave additional insights on how RL mechanisms can be implemented in SNN. Since the SNN proposed has no features explicitly depending on the RL task solved, it is reasonable to hope that the approach developed in this study will be applicable to wide range of RL tasks. It should be confirmed in further research works. Particularly, the next stages of this research project should include:

- application of the approach described to the standard RL benchmarks – to compare it with the state-of-the-art methods;
- extension of my approach to the RL tasks where memory should be a necessary element.

In the RL task described in this article, the "stimulus-reaction" behavior model of the network is quite sufficient, but in the real-world tasks, the network will have to take into account some recent history, some context to issue correct commands. Therefore, a working memory mechanism will be necessary.

Besides that, implementation of the SNN described in this paper on the Loihi neuroprocessor is planned on the next step of this research.

## Acknowledgements.

The present work is a part of the research project in the field of SNN carried out by Chuvash State University in cooperation with Kaspersky and the private company Cifrum. Access to Loihi-based computational systems is provided as a part of Chuvash State University participation in Intel's INCR program.

Cifrum's GPU cluster was used for running the optimization procedure described in Section 3. All other computations were performed on my own computers equipped with GPUs. My SNN emulator package ArNI-X was used to obtain all results reported in this paper.

## References.